\documentclass[10pt]{article}

\usepackage[letterpaper,margin=0.82in]{geometry}
\usepackage{mathptmx}
\usepackage{microtype}
\usepackage{amsmath,amssymb}
\usepackage{booktabs,multirow,tabularx,array}
\usepackage{graphicx}
\usepackage[dvipsnames]{xcolor}
\usepackage{tikz}
\usetikzlibrary{arrows.meta,calc,positioning,shapes.geometric}
\usepackage[numbers,sort&compress]{natbib}
\usepackage[hidelinks]{hyperref}
\usepackage{caption}
\usepackage{titlesec}
\usepackage{float}
\usepackage{placeins}

\makeatletter
\@ifundefined{vcenter@text}{%
  \protected\def\vcenter@text{%
    \check@mathfonts
    \afterassignment\vcenter@text@auxi
    \setbox0\vbox}
  \protected\def\vcenter@text@auxi{\aftergroup\vcenter@text@auxii}
  \protected\def\vcenter@text@auxii{%
    \begingroup
      \dimen0\dimexpr\ht0+\dp0\relax
      \ht0\dimexpr\dimexpr\ifodd\dimen0 1sp + \fi\dimen0\relax/2
        +\fontdimen22 \textfont2\relax
      \dp0\dimexpr\dimen0-\ht0\relax
      \box0
    \endgroup}
}{}
\makeatother

\titleformat{\section}{\large\bfseries}{\thesection}{0.5em}{}
\titleformat{\subsection}{\normalsize\bfseries}{\thesubsection}{0.45em}{}
\titlespacing*{\section}{0pt}{10pt plus 2pt minus 1pt}{4pt}
\titlespacing*{\subsection}{0pt}{7pt plus 1pt minus 1pt}{3pt}
\newcommand{\method}{LocalHK}
\newcommand{\base}{HKTex}
\newcommand{\thermal}{ThermalRF}
\newcommand{\pq}{LocalHK-PQ}
\newcommand{\pf}{LocalHK-PF}
\newcommand{\best}[1]{\textbf{#1}}

\title{\vspace{-1.4em}\textbf{Accelerating HKTex without Mesh Eigensystems:\\Local Unfolding and Randomized Thermal Features}\vspace{-0.3em}}
\author{%
Zhewen He$^{*}$ \qquad Junyi Hu$^{*}$ \qquad Yi Fang\\
New York University Abu Dhabi\\[-0.15em]
{\small $^{*}$Equal contribution}
}
\date{}

\begin{document}
\maketitle
\vspace{-1.4em}

\begin{abstract}
Heat Kernel Textures (\base) represent surface appearance with intrinsic anisotropic kernels, but evaluate them using 50 global Laplace--Beltrami eigendecompositions and a resident basis of shape $[50,V,256]$. We study two complementary ways to remove this bottleneck while leaving the trainer, GeodesicOpt, density control, and compositing unchanged. \method{} exploits the measured locality of trained kernels and replaces spectral evaluation by radius-bounded hinge unfolding and an analytic log-map kernel. On 10 Objaverse meshes and an 8-mesh low-poly holdout, it changes mean view PSNR from 31.35 to 32.00 and from 29.82 to 30.76 dB, respectively, while reducing initialization by 40.5 times and enabling a 749,570-vertex proxy-backed run where the spectral baseline fails. \thermal{} instead preserves the discrete anisotropic heat semigroup: GPU sparse Chebyshev actions and randomized range finding construct global low-rank heat factors without mesh-sized eigenvectors, and a compiled evaluator mixes four neighboring thermal responses. On spot and a thin-stem challenge, \thermal{} reduces end-to-end preprocessing, initialization, and 5,000-step optimization by 29.3\% and 24.4\%, with every surface, atlas, or view PSNR change within 0.12 dB and training allocation reduced by about 90\%. The two routes expose a useful design choice: maximal locality and scale versus fidelity to the thermal PDE. Broader thermal-feature evaluation and real large scenes remain future work.
\end{abstract}

\section{Introduction}

Surface appearance is conventionally stored in a two-dimensional atlas attached to a mesh. Although efficient and well supported, UV maps introduce seams, distortion, duplicated vertices, wasted atlas space, and an authoring problem: a suitable global parameterization must be found before colors can be stored. Alternatives attach color samples directly to mesh elements~\cite{yuksel2010mesh,yuksel2017mesh}, learn continuous texture fields in ambient space~\cite{oechsle2019texture}, or discover neural parameterizations~\cite{xiang2021neutex}. Each changes the trade-off among continuity, editability, model dependence, storage, and evaluation cost.

\base{} takes a different route. Inspired by Gaussian primitives and adaptive density control~\cite{kerbl2023gaussian}, it places anisotropic heat kernels directly on a mesh and optimizes their locations intrinsically~\cite{foti2026heat}. This avoids UV parameterization while retaining an explicit collection of appearance primitives. The cost is a global spectral evaluator: one isotropic and a $7\times7$ grid of anisotropic Laplace--Beltrami operators are eigendecomposed with $k=256$, aligned, and retained as a $[50,V,256]$ basis. The basis alone uses 51.2 kB per vertex.

The implementation is already strongly local after this global step. Each query retrieves 50 candidate kernels in a truncated biharmonic embedding, weights them with a Gaussian window of standard deviation 0.05, and composes only the top 30 responses. This suggests two distinct questions: \emph{can locality replace the heat operator entirely, and can the heat operator itself be applied without computing a global eigensystem?}

For a trained 5,000-kernel spot model, 99\% of contribution lies within radius 0.085 for the median kernel and within 0.216 even in the maximum case. Hard truncation at 0.2 gives 33.623 dB, equal to the untruncated result at the reported precision. This observation motivates \method, a spectral-free evaluator that develops a kernel center only across nearby faces and evaluates an analytic anisotropic response through local log-map coordinates. A geodesic Gaussian window matches the effective shape of \base{} kernels, while a simplified support proxy prevents support cost from growing under subdivision.

Locality is not the only way to remove eigenvectors. Our second route, \thermal, applies each discrete anisotropic heat semigroup directly to random probes using sparse Chebyshev recurrences, then compresses the action into global low-rank factors. Face-interior queries remain barycentric and differentiable. Crucially, learned angle and anisotropy interpolate four heat-kernel \emph{responses}, not aligned eigenvectors. This route retains the thermal PDE at the cost of a global sparse preprocessing pass.

The comparison is intentionally controlled: training samples, GeodesicOpt position updates, pruning, error densification, filtering, top-k color formation, and rendering interfaces remain those of \base. We make five contributions:

\begin{enumerate}
\item We quantify the locality and safe truncation of trained \base{} kernels, and isolate global spectral preprocessing as a scalability bottleneck.
\item We introduce a batched, radius-bounded unfolding evaluator for anisotropic surface kernels, with discrete log maps, filtered support radii, and a tessellation-independent support proxy.
\item We validate the local geometry against exact geodesics, and explain the initial quality gap through the effective kernel shape rather than through distance error or support staleness.
\item We diagnose a low-poly coverage failure, repair it with per-query candidate selection, and evaluate the repair on both the original benchmark and a separately selected holdout.
\item We introduce an eigenvector-free randomized thermal representation and a fused response evaluator, and show on two meshes that it preserves visual quality while accelerating the complete fitting workflow.
\end{enumerate}

\section{Related work}

\subsection{Texture and appearance representations}

Classical texture atlases are compact and directly supported by graphics hardware, but the global UV map couples appearance storage to parameterization quality. Mesh colors attach samples to vertices, edges, and faces to avoid seams~\cite{yuksel2010mesh}; mesh color textures translate this idea into a hardware-filterable representation~\cite{yuksel2017mesh}. Their storage and sampling remain tied to mesh elements and their resolution distribution.

Continuous neural alternatives remove a fixed atlas. Texture Fields regress color as a function in three-dimensional space~\cite{oechsle2019texture}. NeuTex learns a reversible mapping between a volumetric surface and a two-dimensional neural texture~\cite{xiang2021neutex}. Intrinsic Neural Fields use Laplace--Beltrami spectral features to learn functions directly on manifolds~\cite{koestler2022intrinsic}. These methods offer flexible learned priors or view-dependent appearance, whereas \base{} and \method{} fit an explicit, compact set of surface primitives using the same optimizer and density controller.

Point and Gaussian primitives form another lineage. Surface splatting derives anisotropic filtering for irregular point samples~\cite{zwicker2001surface}; differentiable surface splatting enables inverse geometry processing~\cite{wang2019dss}; and 3D Gaussian Splatting combines anisotropic primitives with interleaved optimization and density control~\cite{kerbl2023gaussian}. \base{} transfers this primitive-based view to intrinsic mesh textures. \method{} changes neither the primitive parameters nor their optimization, only how a kernel is evaluated on the surface.

\subsection{Heat diffusion and spectral geometry}

Heat kernels encode multi-scale intrinsic geometry. The Heat Kernel Signature restricts them to the diagonal to obtain stable point descriptors~\cite{sun2009hks}, while Heat-Mapping uses heat kernels and the Laplacian spectrum as global, structure-aware coordinates for robust mesh segmentation~\cite{fang2011heatmapping}. Biharmonic distance reweights Laplace--Beltrami eigenfunctions to obtain a smooth global metric~\cite{lipman2010biharmonic}. Anisotropic Laplace--Beltrami operators align diffusion with principal curvature directions~\cite{andreux2015anisotropic}, the construction inherited by \base.

Matrix-function methods provide an alternative to explicit eigenvectors. Polynomial and Krylov schemes approximate the action of a matrix exponential on a block of vectors without forming the dense exponential~\cite{almohy2011action}; randomized range finding then compresses a numerically low-rank operator from such products~\cite{halko2011finding}. \thermal{} combines these ideas for all 49 anisotropic heat operators and retains only global $O(Vr)$ factors.

Our two routes make different use of locality. \method{} replaces eigenfunction interpolation with local developed coordinates and an analytic kernel. \thermal{} preserves the finite-element heat operator but computes a randomized low-rank action instead of its mesh eigensystem.

\subsection{Geodesics and local coordinates}

Exact polyhedral geodesics can be obtained by the MMP window-propagation algorithm~\cite{mitchell1987discrete}. Fast marching solves the Eikonal equation on triangulated manifolds~\cite{kimmel1998geodesic}; the heat method instead uses prefactorable elliptic systems~\cite{crane2013geodesics}. Vector heat extends diffusion to parallel transport and global log maps~\cite{sharp2019vector}. Discrete exponential maps propagate a local parameterization with a Dijkstra-like traversal~\cite{schmidt2006decal}.

\method{} is closest in spirit to discrete local development: crossing a hinge unfolds the adjacent face, so a straight vector in the development approximates the surface log map. Unlike single-source geometry queries, thousands of moving texture kernels are propagated together, truncated by their current learned supports, and rebuilt during optimization. A QSlim proxy~\cite{garland1997surface} bounds this support representation independently of input tessellation.

\section{Background: the global--local mismatch}

\subsection{\base{} kernel evaluation}

For a Laplace--Beltrami operator with eigenpairs $(\lambda_e,\phi_e)$, a truncated spectral heat kernel is
\begin{equation}
 h_t(p,q)\approx \sum_{e=1}^{K}\exp(-\lambda_e t)\phi_e(p)\phi_e(q).
\end{equation}
\base{} constructs one isotropic operator and 49 anisotropic operators sampled over angle and anisotropy. Their eigenvectors are aligned by assignment and Procrustes transforms so that the four neighboring grid cells can be interpolated for each kernel's learned orientation and anisotropy. With $K=256$, all 50 bases reside on the GPU.

Kernel candidates are found by KNN in a 64-dimensional truncated biharmonic embedding. The candidate response is multiplied by a biharmonic Gaussian window, passed through a learned soft threshold, and the largest 30 filtered contributions are normalized into color. Thus global eigenfunctions determine a response that three later mechanisms---KNN, windowing, and top-k compositing---make local.

\subsection{Two acceleration targets}

\base{} aligns neighboring anisotropic bases with full Procrustes rotations before interpolating eigenvectors and eigenvalues. Such rotations preserve mass orthogonality, but mixing eigenvectors associated with different eigenvalues does not preserve the generalized eigen-equation. On six representative spot operators, the relative residual rises from $1.2\times10^{-5}$--$5.6\times10^{-5}$ for the unaligned eigenbasis to 0.18--0.61 after official alignment. Consequently, exact reproduction of a frozen \base{} checkpoint and strict thermal semantics are different goals.

\method{} prioritizes locality and scalability and deliberately changes the kernel family. \thermal{} instead targets the finite-element semigroup of each grid operator, so it mixes four independently valid thermal responses. This also changes the aligned \base{} response and therefore requires retraining; frozen checkpoint agreement is not used as its quality criterion.

\subsection{Observed locality}

\begin{figure}[H]
\centering
\includegraphics[width=\linewidth]{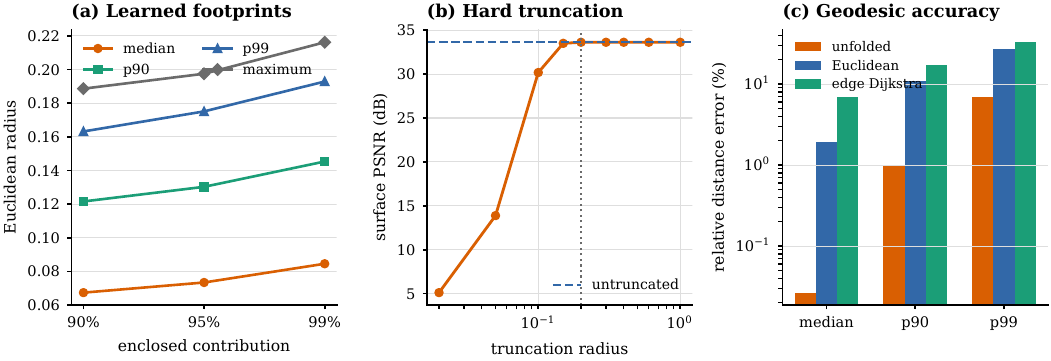}
\caption{Why local evaluation is plausible. (a) Contribution-weighted Euclidean footprint radii for 5,000 trained spot kernels. (b) Surface PSNR after hard response truncation; the dotted line marks $R=0.2$. (c) Relative distance error against MMP exact geodesics.}
\label{fig:locality}
\end{figure}

Figure~\ref{fig:locality} summarizes the motivation. On spot, the contribution-weighted radius grows only modestly as enclosed mass increases from 90\% to 99\%. More directly, setting every response outside $R=0.2$ to zero changes surface PSNR from 33.62314 to 33.62259 dB, while $R=0.1$ drops it to 30.179 dB. Fitting the decay of trained \base{} responses gives an effective diffusion time of 0.00498, far below the nominal 0.0625. The narrow biharmonic window, rather than nominal heat time alone, therefore determines the effective footprint.

\section{Route I: local spectral-free kernels}

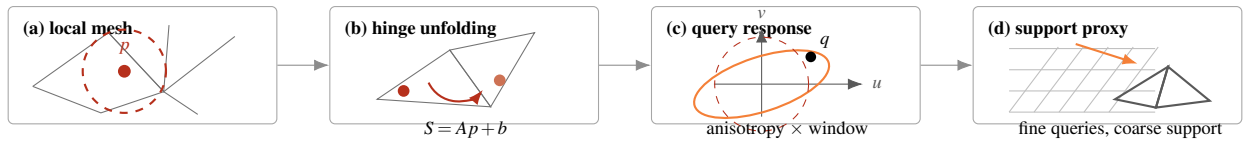
\begin{figure}[H]
\centering
\begin{tikzpicture}[x=1cm,y=1cm,font=\scriptsize,>={Latex[length=2mm]},
  box/.style={draw=black!35,rounded corners=2pt,minimum width=3.55cm,minimum height=1.55cm,inner sep=2pt},
  lab/.style={font=\scriptsize\bfseries,anchor=north west}]
  \node[box] (a) at (0,0) {};
  \node[box] (b) at (4.25,0) {};
  \node[box] (c) at (8.50,0) {};
  \node[box] (d) at (12.75,0) {};
  \node[lab] at ($(a.north west)+(0.08,-0.08)$) {(a) local mesh};
  \draw[black!55] (-1.45,-0.28)--(-0.45,0.43)--(0.28,-0.35)--(1.22,0.38);
  \draw[black!55] (-1.45,-0.28)--(-0.55,-0.63)--(0.28,-0.35)--(0.73,-0.65);
  \draw[black!55] (-0.45,0.43)--(0.28,-0.35)--(0.35,0.53);
  \fill[BrickRed] (-0.24,-0.08) circle (2.4pt);
  \draw[BrickRed,thick,dashed] (-0.24,-0.08) circle (0.55);
  \node[BrickRed,anchor=south] at (-0.24,0.03) {$p$};

  \node[lab] at ($(b.north west)+(0.08,-0.08)$) {(b) hinge unfolding};
  \draw[black!55] (3.10,-0.45)--(4.06,0.20)--(4.60,-0.55)--cycle;
  \draw[black!55] (4.06,0.20)--(5.18,0.43)--(4.60,-0.55)--cycle;
  \draw[BrickRed,thick,->] (3.78,-0.23) arc[start angle=205,end angle=320,radius=0.43];
  \fill[BrickRed] (3.45,-0.34) circle (2.2pt);
  \fill[BrickRed!65] (4.72,-0.20) circle (2.2pt);
  \node[anchor=north] at (4.27,-0.62) {$S=A p+b$};

  \node[lab] at ($(c.north west)+(0.08,-0.08)$) {(c) query response};
  \draw[->,black!65] (7.55,-0.25)--(9.52,-0.25) node[right] {$u$};
  \draw[->,black!65] (8.18,-0.68)--(8.18,0.50) node[above] {$v$};
  \draw[Orange,thick,rotate around={18:(8.18,-0.25)}] (8.18,-0.25) ellipse (0.93 and 0.36);
  \draw[BrickRed,dashed] (8.18,-0.25) circle (0.62);
  \fill[black] (8.83,0.11) circle (2pt) node[anchor=south west] {$q$};
  \node[anchor=north] at (8.52,-0.62) {anisotropy $\times$ window};

  \node[lab] at ($(d.north west)+(0.08,-0.08)$) {(d) support proxy};
  \foreach \x in {11.45,11.78,...,12.78}{\draw[black!25] (\x,-0.62)--(\x+0.65,0.24);}
  \foreach \y in {-0.62,-0.34,-0.06,0.22}{\draw[black!25] (11.45,\y)--(13.38,\y);}
  \draw[->,Orange,thick] (12.33,0.28)--(13.15,0.02);
  \draw[black!65,thick] (12.85,-0.45)--(13.55,-0.02)--(13.39,-0.56)--cycle;
  \draw[black!65,thick] (13.55,-0.02)--(14.10,-0.47)--(13.39,-0.56)--cycle;
  \node[anchor=north] at (12.92,-0.62) {fine queries, coarse support};
  \draw[->,black!45] (a.east)--(b.west);
  \draw[->,black!45] (b.east)--(c.west);
  \draw[->,black!45] (c.east)--(d.west);
\end{tikzpicture}
\caption{\method{} overview. Face-local hinges develop a center into nearby faces; the development defines log-map coordinates and an anisotropic response with a narrow geodesic window. Supports may live on a simplified proxy while colors are queried on the original mesh.}
\label{fig:method}
\end{figure}

\subsection{Face-local precomputation}

Let a triangular mesh have vertices $\mathcal V$ and faces $\mathcal F$. For each face $f$, we precompute its three neighbors, centroid and radius, normal, and a tangent frame $(e_{1f},e_{2f})$. To isolate kernel computation, the first axis follows \base's convention: average the maximum-principal-curvature directions at the face vertices and project into the face plane, falling back to an edge direction only if the projection degenerates.

For adjacent faces, the common edge defines a hinge axis. A Rodrigues rotation and translation rigidly map the plane of the source face onto that of the target. These quantities depend only on the mesh and require storage linear in the face count.

\subsection{Radius-bounded batched development}

All kernel centers expand simultaneously. The state associated with a reached kernel--face pair $(i,f)$ is a rigid map
\begin{equation}
S_{if}=A_{if}p_i+b_{if},
\end{equation}
where $p_i$ is the current center and $S_{if}$ is its development in the plane of $f$. Pairs advance in increasing distance bands, analogous to batched Dijkstra propagation, and each pair is finalized once. This prevents cycles around cone vertices.

A radius test retains a target face when the distance from its centroid to $S_{if}$, minus the face radius, intersects the kernel support. An edge-window test further requires the straight segment from the developed center to the target-face centroid to cross the shared edge, with a small tolerance at its endpoints. This removes many unfolding paths that are locally inconsistent while keeping the traversal branch-free at query time.

\subsection{Log map and kernel response}

For query $q$ on reached face $f$, the developed displacement is rotated back to the source tangent plane:
\begin{equation}
\delta_{if}=A_{if}^{\mathsf T}(q-S_{if}),\qquad
d_{if}=\max\!\left(\lVert q-S_{if}\rVert_2,\lVert q-p_i\rVert_2\right).
\label{eq:logmap}
\end{equation}
The second term ensures that the estimate cannot become shorter than the Euclidean chord. Let $(u,v)$ be the components of $\delta$ along the kernel's learned rotation of the source principal frame. The unfiltered response is
\begin{equation}
x_i(q)=
\exp\!\left[-\frac{d^2}{4t}\frac{u^2+(1+\eta_i)v^2}{u^2+v^2}\right]
\exp\!\left[-\frac{d^2}{2\sigma^2}\right],
\label{eq:kernel}
\end{equation}
with nominal diffusion time $t=0.0625$ and geodesic-window standard deviation $\sigma=0.1$. The first factor is the local tangent-plane analogue of \base's anisotropic heat response. The second reproduces its empirically narrow effective window. The response then passes through the unchanged learned soft threshold and color-formation rule.

\subsection{Support radius, candidate selection, and proxy}

For each kernel, we invert the soft threshold to find the radius at which the filtered response reaches its 1\% tail. Its support is the smaller of this value and 0.2. Supports are rebuilt every 10 optimization steps. All center, angle, anisotropy, threshold, sharpness, and color parameters are concatenated into one table, so a single indexed gather retrieves candidate parameters and uses one indexed-add operation in the backward pass.

The original implementation retained the 50 candidates nearest each face centroid. This is correct when faces are small relative to kernels, but not on a large triangle containing many centers. Our final rule is:
\begin{itemize}
\item if a face supports at most 50 kernels, every query uses all of them;
\item otherwise each query selects its own nearest 50 developed centers using the distance in Eq.~\eqref{eq:logmap}.
\end{itemize}
This recovers the per-query semantics of \base's KNN without adding a new candidate-count hyperparameter.

Finally, if the input is substantially finer than the kernel scale, QSlim constructs a coarse support proxy. Input-face centroids map to proxy faces, and support propagation and candidate tables live on the proxy. Queries and color evaluation remain on the original mesh. The final automatic policy only decimates fine inputs; it does not refine coarse inputs, because proxy refinement was empirically harmful there.

\subsection{Controlled integration and cost}

\method{} replaces the objects that interpolate \base{} eigenbases and evaluate spectral kernels. The data module, optimizer, GeodesicOpt tracer, pruning and error-based densification, soft filter, top-30 compositing, mean-color residual, and rendering interface are unchanged. This makes the comparison narrower than a comparison between independently designed texture systems.

Let $\mathcal P$ be the set of stored kernel--proxy-face pairs. Mesh-local precomputation is linear in proxy size; persistent support storage is $O(|\mathcal P|)$; and query work is bounded by the 50 candidates and top 30 composition. Without a proxy, $|\mathcal P|$ grows under subdivision because a fixed-radius patch intersects more faces. With a fixed-scale proxy, it is governed by kernel count and physical footprint rather than textured-mesh tessellation.

\section{Route II: randomized thermal features}

\subsection{Heat action without a mesh eigensystem}

For anisotropic grid operator $j$, let $L_j$ be the finite-element stiffness matrix, $M$ the lumped mass matrix, and
\begin{equation}
A_j=M^{-1/2}L_jM^{-1/2},\qquad
H_j=M^{-1/2}\exp(-tA_j)M^{-1/2}.
\label{eq:thermal-operator}
\end{equation}
Rather than diagonalizing $A_j$, we approximate the action of $\exp(-tA_j)$ on a block by a Chebyshev recurrence. A Gershgorin bound determines the spectral interval, and coefficients are truncated when the uniform tail is below $10^{-5}$. The implementation uses sparse CSR--dense products on the GPU.

For a Gaussian matrix $\Omega\in\mathbb R^{V\times s}$, we compute
\begin{equation}
Y=\exp(-tA_j)\Omega,\quad Q=\operatorname{qr}(Y),\quad
B=Q^{\mathsf T}\exp(-tA_j)Q.
\end{equation}
Only the small projected matrix is decomposed, $B=U\Sigma U^{\mathsf T}$. The retained factor is
\begin{equation}
F_j=M^{-1/2}QU_r\Sigma_r^{1/2},\qquad H_j\approx F_jF_j^{\mathsf T}.
\label{eq:thermal-factor}
\end{equation}
Our final setting uses rank $r=192$, oversampling 64 ($s=256$), and stores the 49 factors in float16. Thus no $V\times K$ mesh eigenvectors are computed; the only eigendecomposition is $256\times256$.

\subsection{Barycentric responses and parameter interpolation}

For a point $p$ with triangle vertices $(v_1,v_2,v_3)$ and barycentric coordinates $b$, factor interpolation is linear,
\begin{equation}
\bar F_j(p)=\sum_{a=1}^{3}b_aF_j(v_a).
\end{equation}
If kernel $i$ lies between four angle--anisotropy grid operators $j_c$ with bilinear weights $w_{ic}$, its response is
\begin{equation}
h_i(p,q)=\sum_{c=1}^{4}w_{ic}\,\bar F_{j_c}(p)^{\mathsf T}\bar F_{j_c}(q).
\label{eq:response-mixture}
\end{equation}
Equation~\eqref{eq:response-mixture} mixes heat-kernel responses rather than eigenvectors. Every grid term is a positive-semidefinite low-rank approximation to Eq.~\eqref{eq:thermal-operator}, and gradients pass through barycentric coordinates and grid weights. Self heat $h_i(p,p)$ provides the same normalization interface as \base.

The isotropic eigensystem used for candidate routing is also removed. We sample a randomized heat-filtered biharmonic embedding
\begin{equation}
Z=M^{-1/2}A_0^{+}\exp(-\tau A_0)G/\sqrt d,
\end{equation}
using one nullspace-constrained sparse Poisson factorization, with $d=128$ and $\tau=0.0025$. FAISS still selects 50 candidates; the original Gaussian post-diffusion window, learned soft threshold, top-30 composition, and all optimization rules remain unchanged.

\subsection{Fused implementation}

A direct implementation performs 12 feature gathers (four operators times three triangle vertices) and four rank-$r$ dot products per query--candidate pair. We fuse these operations with dynamic \texttt{torch.compile} kernels while retaining the explicit four-response sum. This changes neither Eq.~\eqref{eq:response-mixture} nor its gradients, but avoids materializing the full corner-by-rank query tensor. The one-time compilation cost is included in all reported optimization wall times.

\section{Experimental setup}

\subsection{Data and configurations}

We use Keenan Crane's spot mesh for controlled locality, geometry, ablation, and subdivision studies. It has 2,930 vertices and 5,856 faces after loading, and is normalized to maximum bounding-box extent 2.

For the main quality benchmark, we reuse the \base{} authors' Objaverse~\cite{deitke2023objaverse} eligibility logic: downloadable and textured, fewer than 60,000 vertices, static, manifold, one connected component, limited ray self-intersections, and successful $k=256$ eigendecomposition. Meshes used by the authors for tuning are excluded. A fixed seed selects 10 meshes. Each method is planned for three optimization seeds.

The low-poly holdout is selected only after the per-face failure has been diagnosed. It applies the same eligibility conditions, adds fewer than 3,000 annotated vertices, uses a different selection seed, excludes the main set, and contains 8 meshes. It therefore tests whether the per-query repair transfers to unseen low-poly geometry, though it is not a fully independent benchmark.

We compare \base, \pf{} (windowed \method{} with per-face candidates), and \pq{} (the final per-query method). All use the \base{} learning rates, parameter constraints, GeodesicOpt, importance sampling, pruning, and densification settings. The scaling study predates the window and per-query repair and is explicitly labeled V0.

The \thermal{} feasibility study uses two 5,000-step, seed-0 pairs: spot and Objaverse \texttt{06d17021}. The latter is the pumpkin whose thin stem and sharp attachment expose \method's remaining coverage failure. Each pair runs on one A100 node, forces a fresh \base{} spectral cache, and evaluates rank-192/oversampling-64 thermal factors with 128-dimensional randomized routing. Rank and routing dimension are selected on spot; the pumpkin is a transfer case.

\subsection{Evaluation protocol}

Both methods are wrapped as a color function of face index and barycentric coordinates and are queried on identical points. We report:
\begin{itemize}
\item surface PSNR on 200,000 area-uniform samples;
\item PSNR, masked SSIM, and LPIPS on a $1{,}024^2$ UV rasterization;
\item foreground PSNR, SSIM, and LPIPS on 8 rasterized albedo views at $512^2$.
\end{itemize}
This evaluator does not reproduce the rotating Mitsuba protocol in the \base{} paper. It renders albedo only and should not be called the official \base{} metric. The two \thermal{} pairs additionally report the paper implementation's five-frame rotating Mitsuba PSNR, SSIM, MS-SSIM, and LPIPS.

For multi-seed results, we first average available seeds for each mesh and then give every mesh equal weight. Paired standard deviations are sample standard deviations across meshes. Exact one-sided sign tests are exploratory because of small samples and post-hoc method development.

\subsection{Timing and completion}

Main experiments ran on shared, heterogeneous A100 GPUs. Initialization includes method-specific preprocessing; optimization time includes support rebuilding and density-control events; step time is the median after warm-up; peak training memory is allocated GPU memory during optimization. Query throughput uses the shared 200,000-point evaluator. These measurements are noisy but use the same instrumentation for both methods.

For \thermal{}, end-to-end time is defined explicitly. The \base{} total is forced-fresh trainer initialization plus optimization. The \thermal{} total is offline factor and routing construction plus trainer initialization plus optimization; it includes first-shape compilation. Steady view-query throughput sums foreground pixels and texture-query time over the eight views, avoiding the one-time compilation triggered by the first 8,192-point evaluation shape.

\base{} completes 30/30 main runs; \pf{} and \pq{} each complete 28/30. The missing \method{} runs occur on a mesh where the shared DiGeo tracer raises CUDA errors. The holdout completes 68/72 runs across three methods; four unfinished runs were canceled at stage end. Holdout timing is excluded because later jobs packed four experiments on one GPU.

\section{Results}

\subsection{Locality and geometric validation}

Figure~\ref{fig:locality}(a--b) establishes that the useful \base{} response is compact despite its global representation. The 90\%-contribution radius is 0.067 at the median and 0.163 at p99. For 99\% contribution, the median is 0.085, p99 is 0.193, and the maximum is 0.216. The hard-truncation curve remains flat from radius 0.2 onward.

For geometry, we sample 40 sources, propagate to radius 0.4, and compare 8,108 source-query pairs with MMP exact geodesics. Table~\ref{tab:geometry} and Fig.~\ref{fig:locality}(c) show that unfolding is substantially more accurate than both Euclidean chord distance and edge-graph Dijkstra. Its direction differs from a vector-heat log map by 1.9 degrees at the median and 11.0 degrees at p90. The edge-window test misses 2.8\% of exact-radius pairs; support propagation for 5,000 kernels at radius 0.15 takes 0.575 s on CPU. The later texture ablation asks whether this geometric improvement matters to appearance on smooth spot.

\begin{table}[H]
\centering
\caption{Spot geodesic validation against MMP exact distance. Values are relative errors in percent over 8,108 pairs.}
\label{tab:geometry}
\begin{tabular}{lrrr}
\toprule
Distance estimate & Median & p90 & p99 \\
\midrule
Local unfolding & \best{0.03} & \best{1.0} & \best{6.9} \\
Euclidean chord & 1.9 & 10.8 & 27.4 \\
Edge-graph Dijkstra & 6.9 & 17.1 & 32.9 \\
\bottomrule
\end{tabular}
\end{table}

\subsection{Aggregate and per-mesh quality}

Table~\ref{tab:quality} shows that \pq{} slightly improves every aggregate quality metric in both sets. On the main set its paired view-PSNR change is $+0.647\pm1.444$ dB, with 9/10 mesh wins; LPIPS improves on 10/10. The exploratory one-sided sign test is $p=0.010742$. On the untouched low-poly holdout, view PSNR changes by $+0.945\pm0.717$ dB, with 7/8 wins ($p=0.035156$); LPIPS improves on 8/8.

\begin{table}[H]
\centering
\caption{Quality under our custom albedo evaluation. PF uses one candidate set per face; PQ selects candidates per query on overflowing faces. Means weight meshes equally.}
\label{tab:quality}
\begin{tabular}{llrrrr}
\toprule
Set & Method & View PSNR$\uparrow$ & SSIM$\uparrow$ & LPIPS$\downarrow$ & Surface PSNR$\uparrow$ \\
\midrule
\multirow{3}{*}{Main (10)}
 & \base & 31.35 & 0.937 & 0.065 & 31.68 \\
 & \pf   & 29.37 & 0.935 & 0.072 & 29.72 \\
 & \pq   & \best{32.00} & \best{0.940} & \best{0.058} & \best{32.39} \\
\midrule
\multirow{3}{*}{Holdout (8)}
 & \base & 29.82 & 0.947 & 0.058 & 29.65 \\
 & \pf   & 29.92 & 0.946 & 0.054 & 29.62 \\
 & \pq   & \best{30.76} & \best{0.954} & \best{0.050} & \best{30.76} \\
\bottomrule
\end{tabular}
\end{table}

\begin{figure}[H]
\centering
\includegraphics[width=\linewidth]{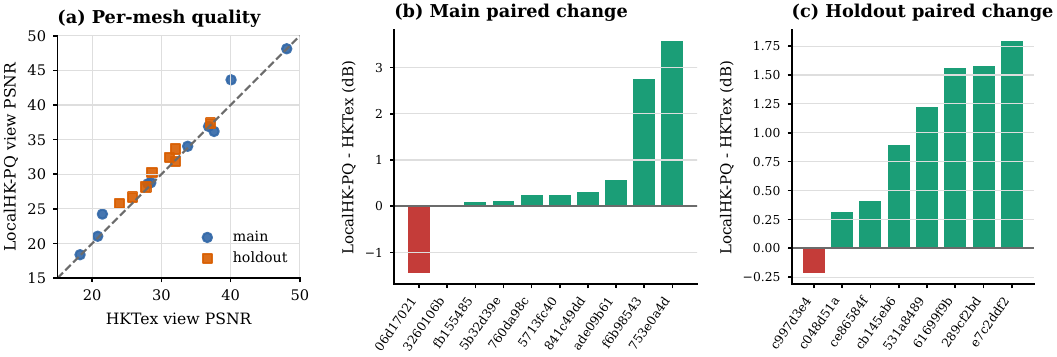}
\caption{Per-mesh view PSNR. (a) Final \pq{} versus \base; the dashed line is equality. (b--c) Paired changes for the main and low-poly holdout sets. Mesh identifiers are shortened to eight characters.}
\label{fig:quality}
\end{figure}

The scatter in Fig.~\ref{fig:quality} separates average improvement from variation. Most points remain close to the equality line, but the main set contains a large positive outlier and one negative pumpkin case. The holdout has smaller spread and one slight negative. Full per-mesh numbers appear in Appendix~\ref{app:permesh}.

\subsection{Efficiency and scaling}

Table~\ref{tab:efficiency} separates different notions of speed. \pq{} reduces initialization by 40.5 times, peak training memory by 8.22 times, and raises texture-query throughput by 9.97 times. Its median step is 1.53 times faster. However, optimization wall time is effectively unchanged (141.83 versus 143.97 s) because periodic support rebuilds and density-control steps dominate parts of the run.

\begin{table}[H]
\centering
\caption{Main-set efficiency. Gain is \base/\pq{} for time and memory, and \pq/\base{} for throughput. Optimization wall time is not presented as a speedup.}
\label{tab:efficiency}
\begin{tabular}{lrrrr}
\toprule
Metric & \base & \pf & \pq & Gain \\
\midrule
Initialization (s)       & 163.75 & 3.83 & 4.04 & 40.5$\times$ \\
Median step (ms)         & 23.76  & 12.87 & 15.56 & 1.53$\times$ \\
Optimization (s)         & 141.83 & 135.98 & 143.97 & --- \\
Peak train memory (GB)   & 2.952  & 0.284 & 0.359 & 8.22$\times$ \\
Query rate (M points/s)  & 0.145  & 3.670 & 1.443 & 9.97$\times$ \\
\bottomrule
\end{tabular}
\end{table}

\begin{figure}[H]
\centering
\includegraphics[width=\linewidth]{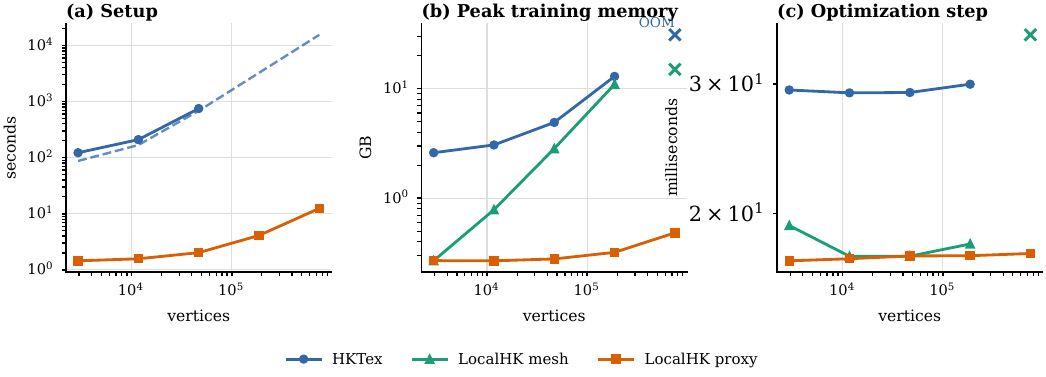}
\caption{Scaling on midpoint-subdivided spot with 5,000 kernels and 120 steps. Full 50-eigensolve measurements stop at 46,850 vertices; the dashed setup curve extrapolates 49 anisotropic solves from one measured solve. Crosses mark OOM rather than measured values. This experiment uses early V0 \method{} without the geodesic window or per-query selection.}
\label{fig:scaling}
\end{figure}

The subdivision sequence in Fig.~\ref{fig:scaling} holds spot geometry and texture fixed while increasing tessellation. Full 50-eigensolve preprocessing grows from 122 s at 2,930 vertices to 750 s at 46,850. Extrapolating 49 anisotropic solves from one measured solve gives lower bounds of 3,248 s (about 54 min) at 187,394 vertices and 15,471 s (about 4.3 h) at 749,570. The largest basis alone is 38.38 GB. In our shared 48-GB A6000 run, \base{} fails during initialization while attempting a 35.74-GiB allocation.

Without a proxy, \method{} support storage also grows with subdivision and fails during optimization at 749,570 vertices. With the proxy, memory changes only from 0.269 to 0.483 GB, median step from 17.23 to 17.63 ms, and query throughput from 3.393 to 3.443 million points/s between the smallest and largest meshes. The input itself still costs memory and setup time, but support complexity is effectively fixed. Across the 10 benchmark meshes, full \base{} spectral preprocessing totals 2,454.7 s; even the 121-vertex mesh takes 17.4 s.

\subsection{Thermal-preserving acceleration}

Table~\ref{tab:thermal} reports the complete \thermal{} pairs. On spot, the final representation changes surface/atlas/view PSNR by $+0.043/-0.117/-0.055$ dB and paper PSNR by $-0.001$ dB. On the pumpkin transfer case, the corresponding changes are $+0.068/+0.144/+0.016$ dB and $-0.056$ dB. View LPIPS improves by 0.00100 and 0.00139, respectively. Direct inspection of both saved atlases and difficult views finds no new seam, structured blur, or thin-stem artifact.

\begin{table}[H]
\centering
\caption{Thermal-preserving 5,000-step pairs. E2E includes fresh preprocessing, initialization, first-shape compilation, and optimization. Paper is five-frame rotating Mitsuba PSNR.}
\label{tab:thermal}
\small
\setlength{\tabcolsep}{3.2pt}
\begin{tabular}{llrrrrrrr}
\toprule
Mesh & Method & Surface & Atlas & View & Paper & Step (ms) & E2E (s) & Train GB \\
\midrule
\multirow{2}{*}{spot}
 & \base     & 33.859 & 33.659 & 32.961 & 41.978 & 28.88 & 222.63 & 2.599 \\
 & \thermal  & 33.902 & 33.542 & 32.906 & 41.977 & \best{21.54} & \best{157.34} & \best{0.257} \\
\midrule
\multirow{2}{*}{pumpkin}
 & \base     & 37.265 & 36.844 & 37.591 & 41.349 & 28.97 & 244.71 & 2.598 \\
 & \thermal  & 37.333 & 36.989 & 37.607 & 41.294 & \best{21.83} & \best{185.11} & \best{0.257} \\
\bottomrule
\end{tabular}
\end{table}

The low-rank approximation itself is measured on all 49 spot operators. At rank 256, the worst/median filtered-response PSNR against the unaligned 256-mode thermal oracle is 62.15/78.41 dB, with 1.46\% maximum raw relative error. The faster rank-192 setting reaches 40.30/52.08 dB and 6.49\% maximum raw error. Despite the looser operator approximation, retraining absorbs the residual without measurable appearance loss in these pairs.

On the same nodes, \thermal{} reduces median step time by 25.4\% and 24.6\%, end-to-end time by 29.3\% and 24.4\%, and peak training allocation by about 90\%. Its factors plus routing occupy about 56.5 MB on each mesh, versus roughly 153 MB of resident \base{} geometry tensors. Warm view-query throughput improves from 0.452 to 0.990 million points/s on spot and from 0.450 to 1.009 million points/s on pumpkin. Summed across both pairs, end-to-end time falls from 467.34 to 342.45 s, a 1.365-times speedup.

\subsection{What closes the spot quality gap?}

Initial \method{} underperforms \base{} on spot by 2.45 dB surface PSNR and visibly softens boundaries. Table~\ref{tab:ablation} rules out several implementation hypotheses. Doubling the candidate cap, using Euclidean geometry, rebuilding supports every step, or reducing the hard radius to 0.2 changes V0 by at most 0.26 dB. Removing anisotropy, however, loses 7.94 dB: the directional kernel family is essential.

\begin{figure}[H]
\centering
\includegraphics[width=\linewidth]{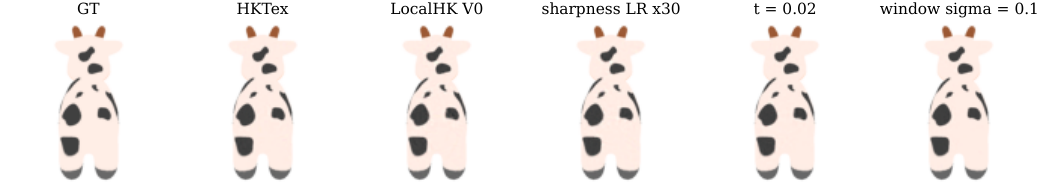}
\caption{Spot view for the kernel-shape investigation. Anisotropic V0 is too broad; optimization changes help, while the geodesic Gaussian window most closely restores sharp boundaries. These are single-seed diagnostic runs.}
\label{fig:spotshape}
\end{figure}

The remaining gap is kernel shape. Increasing the sharpness learning rate by 30 times reaches 32.16 dB and reducing nominal $t$ to 0.02 reaches 32.77 dB. Directly initializing at the fitted $t_{\mathrm{eff}}\approx0.005$ is unstable: 65--98\% of samples become uncovered in short diagnostic runs. Keeping the broad nominal heat kernel and multiplying by a fixed geodesic window instead reaches 35.33 dB, 1.71 dB above the spot \base{} run. Figure~\ref{fig:spotshape} shows the corresponding boundary sharpening.

\begin{table}[H]
\centering
\caption{Spot surface-PSNR ablation (one mesh, one seed, 5,000 steps). Shape repairs use the radius-0.2 setting; other rows change one V0 component.}
\label{tab:ablation}
\begin{tabular}{lrr}
\toprule
Setting & Surface PSNR & Change from V0 \\
\midrule
\base{} reference & 33.62 & +2.45 \\
\midrule
\method{} V0 & 31.17 & 0.00 \\
isotropic & 23.23 & -7.94 \\
Euclidean distance and frame & 31.02 & -0.15 \\
candidate cap $50\to100$ & 30.97 & -0.20 \\
rebuild every step & 30.91 & -0.26 \\
hard radius 0.2 & 31.29 & +0.12 \\
\midrule
sharpness LR $\times30$ & 32.16 & +0.99 \\
$t=0.02$ & 32.77 & +1.60 \\
window $\sigma=0.1$ & \best{35.33} & \best{+4.16} \\
\bottomrule
\end{tabular}
\end{table}

\subsection{Low-poly coverage failure and repair}

Color residuals are normalized by $\max(\sum_i w_i,1)$. When the filtered weight sum is below one, the prediction is pulled toward the learned mean color, producing gray regions. On the 121- and 330-vertex meshes, a sampled face supports a median of 2,601 and 1,012 kernels. Keeping only the 50 centers nearest the face centroid leaves 36.5\% and 24.6\% of surface samples below weight sum 0.5. Re-evaluating the same kernels without the cap reduces these fractions to 1.2\% and 1.7\%. The cap alone accounts for 35.4\% and 23.0\% of the surface; support propagation and the window account for only 0.35\% and 0.62\%.

\begin{figure}[H]
\centering
\includegraphics[width=\linewidth]{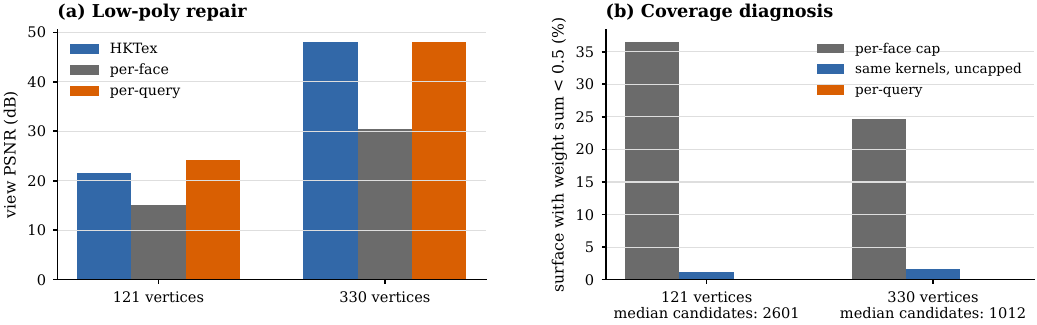}
\caption{The low-poly failure is candidate selection rather than support propagation. Per-query selection restores view PSNR (left) by eliminating the large under-covered regions created by the per-face cap (right).}
\label{fig:coverage}
\end{figure}

Per-query selection eliminates the diagnostic coverage hole and recovers quality (Table~\ref{tab:coverage}, Fig.~\ref{fig:coverage}). On the 330-vertex mesh it matches \base{}; on the 121-vertex mesh it surpasses \base{} by 2.76 dB. The repair is not free: the main-set mean step increases from 12.87 to 15.56 ms and query throughput decreases from 3.670 to 1.443 million points/s. On the 121-vertex mesh, \pq{} takes 34.6 ms per step versus 20.6 ms for \base{} on the same mesh, and is also slower than the 23.8-ms main \base{} mean.

\begin{table}[H]
\centering
\caption{Coverage diagnosis and view PSNR on the two coarse meshes. Coverage statistics use the final kernels of a single diagnostic seed; PSNR averages available benchmark seeds.}
\label{tab:coverage}
\begin{tabular}{lrrrrrrr}
\toprule
Vertices & Cand. med. & PF $<0.5$ & Uncapped $<0.5$ & Cap-only & \base & \pf & \pq \\
\midrule
121 & 2,601 & 36.5\% & 1.2\% & 35.4\% & 21.49 & 15.03 & \best{24.25} \\
330 & 1,012 & 24.6\% & 1.7\% & 23.0\% & 48.11 & 30.52 & \best{48.13} \\
\bottomrule
\end{tabular}
\end{table}

\subsection{Qualitative cases and remaining failure}

\begin{figure}[H]
\centering
\includegraphics[width=\linewidth]{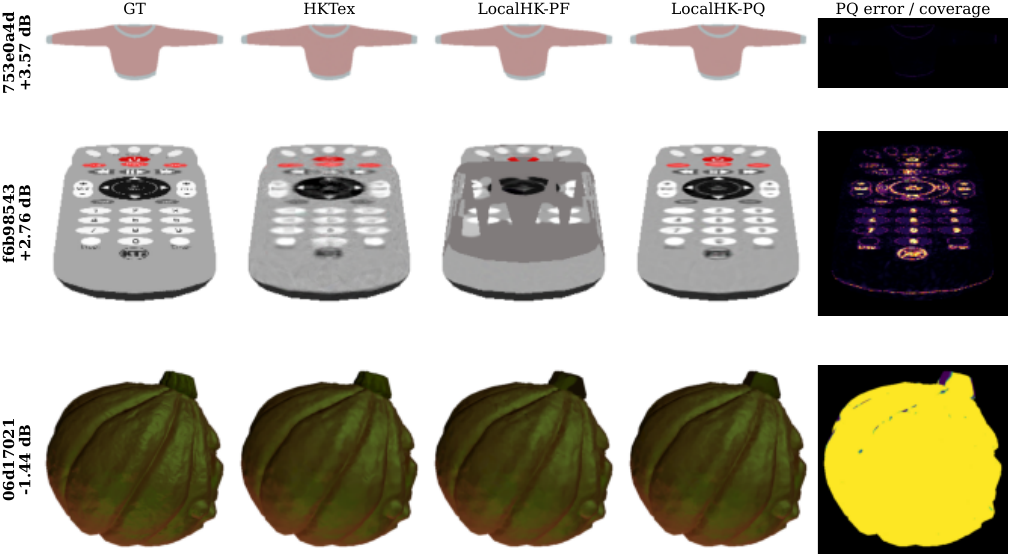}
\caption{Representative main-set views. Rows show a positive case, the repaired low-poly case, and the remaining pumpkin failure. The last column is \pq{} error for the first two rows and coverage for the pumpkin (yellow covered, purple uncovered). Labels give paired view-PSNR changes.}
\label{fig:gallery}
\end{figure}

Figure~\ref{fig:gallery} shows three distinct cases. The sweater is the largest positive main-set change; the initial per-face method is already strong and per-query selection preserves it. The low-poly remote control exposes the gray under-coverage and its repair. The pumpkin is the only main-set mesh on which final \method{} remains materially below \base{}: 36.19 versus 37.62 dB, a -1.44 dB gap.

The pumpkin coverage map localizes the problem to the stem and a few thin grooves. Removing the per-face cap does not change coverage, so this is not the repaired failure. Kernels on the body do not propagate through the sharp attachment into the thin stem, while error densification only splits existing kernels and cannot seed an empty region. A natural extension is coverage-driven insertion at high-error points with low weight sum.

\section{Discussion}

\paragraph{What is actually gained?}
The strongest result is not a small quality change; it is that both routes remove the mesh eigensystem without changing the rest of the texture learner. \method{} eliminates the heat solve itself and exposes strong locality and proxy scaling. \thermal{} is a more conservative numerical replacement: it retains a globally supported approximation of each anisotropic heat semigroup and still reduces complete fitting time on both tested meshes.

\paragraph{Which route should be used?}
The answer depends on the contract. When proxy-backed scaling, very low query cost, or local geometry updates dominate, \method{} is the stronger design. When the anisotropic thermal PDE is semantically required, \thermal{} is the appropriate route: it uses a global sparse pass but no mesh eigenvectors, and its response interpolation remains thermal. The two methods are complementary rather than successive versions of one approximation.

\paragraph{Why not preserve the frozen \base{} output?}
Official Procrustes rotations mix modes with unequal eigenvalues, so the aligned basis is not an eigensystem of its labeled anisotropic operator. A true response-space thermal mixture therefore cannot reproduce a checkpoint trained against that hybrid construction. Our frozen swap loses 6.7 dB on spot, whereas retrained \thermal{} matches \base{}; semantic fidelity must be evaluated after retraining.

\paragraph{Why unfolding rather than Euclidean distance?}
On smooth spot, Euclidean geometry is only 0.15 dB below unfolding even though its geodesic error is much larger. This does not make unfolding unnecessary: Euclidean distance can couple nearby sheets across ambient space and gives no intrinsic direction across folds. Our current benchmark does not isolate thin surfaces, sharp folds, or close disconnected parts, so the texture-level benefit remains unresolved. The geometry validation shows that the local construction is accurate; a targeted dataset is still required to show when that accuracy matters.

\paragraph{What limits scale?}
The proxy removes dependence of support storage on tessellation, not all dependence on input size. Reading geometry, mapping fine faces, sampling, rasterization, and storing the original mesh still scale with $V$ and $F$. Moreover, the largest test is a subdivision of one object rather than a semantically complex scene. The correct conclusion is that the spectral and support bottlenecks are removed in this controlled sequence, not that arbitrary million-face scenes are solved.

\paragraph{Potential extensions.}
For \method{}, coverage-driven seeding could address thin empty components; spatial buckets inside very large faces could accelerate per-query selection; component-local propagation should naturally support disconnected scenes; and local mesh edits should only invalidate nearby data. For \thermal{}, multi-mesh validation should precede rank tuning, while larger original meshes need measured full-grid GPU builds and possibly multilevel or rational actions when Chebyshev degree grows under refinement.

\section{Limitations}

The aggregate \method{} protocol is custom and omits the rotating Mitsuba evaluation of \base. Only 10 main and 8 holdout meshes are tested; the main paired improvement has 1.444 dB mesh-to-mesh standard deviation, and all spot analyses use one seed. The Gaussian window width is selected on spot, and per-query selection is designed after observing main-set failures. The holdout reduces but cannot eliminate this bias. All learning rates and densification thresholds were tuned for \base{}, not \method.

The \thermal{} study is narrower: two meshes, one seed each, and rank/routing choices selected on spot. The pumpkin is a meaningful thin-feature transfer case but not a statistical benchmark. Its five-frame paper metrics are included, yet neither pair establishes quality across the main or holdout sets. The projected factors approximate the full discrete semigroup but are assessed against a 256-mode oracle, which itself is truncated. Reported cold 200,000-point throughput includes compilation for a new batch shape; we therefore report subsequent view-query throughput separately.

The scaling study uses early V0 on a spot subdivision sequence. The final configuration is not retested at 749,570 vertices, and no real large, multi-object, disconnected, or geometry-editing workload is measured. Timing mixes shared A100 variants; the windowed spot run is contended, and holdout timing is unusable because later jobs packed four runs per GPU.

Two \method{} main runs fail in a DiGeo CUDA tracer shared by both methods. Upstream importance-sampling initialization also exits early for both, preserving comparison symmetry but limiting both results. Finally, the eligibility rules explicitly exclude disconnected and spectral-failure meshes. Such inputs may favor a local method, but this study cannot claim that advantage.

\section{Conclusion}

We present two eigenvector-free routes for accelerating \base. \method{} replaces the spectral evaluator by radius-bounded local development; its geodesic window and per-query candidates match or improve average quality across the tested main and holdout sets, while a support proxy reaches 749,570 vertices. \thermal{} instead preserves anisotropic finite-element heat diffusion through GPU polynomial actions and randomized global factors. On spot and a thin-stem transfer case it reduces end-to-end time by 24--29\% and training allocation by about 90\%, with no material quality loss. These results establish two distinct technical options---maximally local analytic kernels and thermal-faithful randomized kernels---rather than one universal replacement. Broader thermal evaluation, official-protocol LocalHK tests, real large scenes, and geometry-editing studies remain necessary for a general claim.

\appendix
\section{Per-mesh quality}
\label{app:permesh}

\begin{table}[H]
\centering
\caption{Main benchmark view PSNR by mesh. Values average available seeds.}
\label{tab:main-permesh}
\small
\begin{tabular}{lrrrrr}
\toprule
Mesh & Vertices & \base & \pf & \pq & $\Delta$(PQ--HK) \\
\midrule
5713fc40 & 56,434 & 33.80 & 34.06 & 34.04 & +0.24 \\
3260106b & 330    & 48.11 & 30.52 & 48.13 & +0.02 \\
fb155485 & 11,224 & 36.82 & 36.90 & 36.91 & +0.09 \\
ade09b61 & 4,767  & 28.05 & 28.62 & 28.62 & +0.57 \\
753e0a4d & 8,138  & 40.08 & 44.28 & 43.64 & +3.57 \\
760da98c & 16,024 & 20.81 & 21.03 & 21.05 & +0.24 \\
06d17021 & 2,912  & 37.62 & 35.97 & 36.19 & -1.44 \\
841c49dd & 15,250 & 28.48 & 28.91 & 28.78 & +0.30 \\
f6b98543 & 121    & 21.49 & 15.03 & 24.25 & +2.76 \\
5b32d39e & 1,495  & 18.26 & 18.38 & 18.38 & +0.12 \\
\midrule
Mean & 11,670 & 31.35 & 29.37 & 32.00 & +0.65 \\
\bottomrule
\end{tabular}
\end{table}

\begin{table}[H]
\centering
\caption{Separately selected low-poly holdout view PSNR by mesh. The last mesh has one completed \pq{} seed and two completed \pf{} seeds; other cells average all available runs.}
\label{tab:holdout-permesh}
\small
\begin{tabular}{lrrrrr}
\toprule
Mesh & Vertices & \base & \pf & \pq & $\Delta$(PQ--HK) \\
\midrule
289cf2bd & 94    & 32.08 & 31.50 & 33.66 & +1.58 \\
e7c2ddf2 & 130   & 23.96 & 21.31 & 25.76 & +1.79 \\
61699f9b & 162   & 28.66 & 29.18 & 30.22 & +1.56 \\
c048d51a & 362   & 37.06 & 37.30 & 37.38 & +0.32 \\
ce86584f & 609   & 27.76 & 28.14 & 28.16 & +0.40 \\
cb145eb6 & 809   & 25.80 & 26.26 & 26.69 & +0.89 \\
531a8489 & 1,060 & 31.16 & 32.46 & 32.39 & +1.23 \\
c997d3e4 & 1,477 & 32.04 & 33.18 & 31.83 & -0.21 \\
\midrule
Mean & 588 & 29.82 & 29.92 & 30.76 & +0.94 \\
\bottomrule
\end{tabular}
\end{table}

\section{Spectral preprocessing by benchmark mesh}

\begin{table}[H]
\centering
\caption{Measured \base{} preprocessing for 50 eigendecompositions on the 10 main meshes. The smallest mesh still pays the fixed 50-solve construction.}
\label{tab:spectral-permesh}
\small
\begin{tabular}{lrrr}
\toprule
Mesh & Vertices & Faces & Spectral preprocessing (s) \\
\midrule
5713fc40 & 56,434 & 112,864 & 1,052.8 \\
760da98c & 16,024 & 32,044  & 333.2 \\
841c49dd & 15,250 & 30,282  & 251.8 \\
fb155485 & 11,224 & 22,448  & 257.9 \\
753e0a4d & 8,138  & 15,976  & 160.6 \\
ade09b61 & 4,767  & 9,530   & 132.2 \\
06d17021 & 2,912  & 5,828   & 116.5 \\
5b32d39e & 1,495  & 2,948   & 92.4 \\
3260106b & 330    & 628     & 39.8 \\
f6b98543 & 121    & 238     & 17.4 \\
\midrule
Total & 116,695 & 232,786 & 2,454.7 \\
\bottomrule
\end{tabular}
\end{table}

\section{Per-mesh efficiency}

\begin{table}[H]
\centering
\caption{Main benchmark efficiency by mesh. Each entry is \base/\pq{} after averaging available seeds. Query rate is in million points/s.}
\label{tab:eff-permesh}
\footnotesize
\begin{tabular}{lccccc}
\toprule
Mesh & Init. (s) & Step (ms) & Memory (GB) & Query (M/s) & Optim. (s) \\
\midrule
5713fc40 & 933.04/6.01 & 23.94/11.93 & 5.832/0.283 & 0.126/1.510 & 149.81/107.20 \\
3260106b & 11.76/2.69  & 23.07/15.40 & 2.409/0.340 & 0.098/0.468 & 145.38/172.08 \\
fb155485 & 142.54/3.35 & 24.08/14.30 & 3.031/0.397 & 0.125/2.070 & 143.60/181.30 \\
ade09b61 & 52.55/3.57  & 23.81/13.18 & 2.675/0.458 & 0.164/1.521 & 134.64/113.28 \\
753e0a4d & 73.05/2.99  & 24.19/11.92 & 2.839/0.224 & 0.127/0.950 & 143.98/120.99 \\
760da98c & 186.36/4.17 & 24.77/12.31 & 3.268/0.654 & 0.122/2.177 & 158.61/131.04 \\
06d17021 & 41.16/2.70  & 23.71/14.38 & 2.577/0.127 & 0.168/2.099 & 140.29/132.77 \\
841c49dd & 160.06/7.93 & 24.81/15.69 & 3.227/0.492 & 0.183/2.327 & 135.82/157.15 \\
f6b98543 & 3.90/4.34   & 20.60/34.65 & 1.164/0.419 & 0.175/0.150 & 122.66/205.28 \\
5b32d39e & 33.10/2.69  & 24.58/11.88 & 2.501/0.198 & 0.158/1.159 & 143.51/118.63 \\
\bottomrule
\end{tabular}
\end{table}

\clearpage
\small
\setlength{\bibsep}{2pt}
\bibliographystyle{abbrvnat}
\bibliography{refs}

\end{document}